\documentclass{article}
\usepackage{spconf,amsmath,graphicx,hyperref}

\usepackage{booktabs}
\usepackage{multirow}
\usepackage{array}
\usepackage{adjustbox}
\usepackage{enumitem}

\usepackage[normalem]{ulem}   
\newcommand{\best}[1]{\textbf{#1}}  
\newcommand{\second}[1]{\uline{#1}}  

\title{MI-Fuse: Label Fusion for Unsupervised Domain Adaptation with Closed-Source Large Audio-Language Model}
\name{Hsiao-Ying Huang*, Yi-Cheng Lin*, Hung-yi Lee \thanks{*Equal Contribution}}
\address{National Taiwan University, Taiwan}
\begin{document}
\ninept
\maketitle
\begin{abstract}
Large audio–language models (LALMs) show strong zero-shot ability on speech tasks, suggesting promise for speech emotion recognition (SER). However, SER in real-world deployments often fails under domain mismatch, where source data are unavailable and powerful LALMs are accessible only through an API. We ask: given only unlabeled target-domain audio and an API-only LALM, can a student model be adapted to outperform the LALM in the target domain? To this end, we propose MI-Fuse, a denoised label fusion framework that supplements the LALM with a source-domain trained SER classifier as an auxiliary teacher. The framework draws multiple stochastic predictions from both teachers, weights their mean distributions by mutual-information-based uncertainty, and stabilizes training with an exponential moving average teacher. Experiments across three public emotion datasets and six cross-domain transfers show consistent gains, with the student surpassing the LALM and outperforming the strongest baseline by 3.9\%. This approach strengthens emotion-aware speech systems without sharing source data, enabling realistic adaptation.
\end{abstract}
\vspace{-8pt}
\section{Introduction}
\vspace{-5pt}
\label{sec:intro}

LALMs such as Desta2.5-Audio \cite{desta2.5_audio}, Qwen2.5-Omni \cite{qwen2.5_omni}, and Gemini 2.5 \cite{gemini2.5} have recently demonstrated impressive general-purpose capabilities across spoken language understanding, paralinguistics, and speaker-related tasks \cite{dynamic_superb_phase_2, wu2024towards}. 
Their versatility and strong zero-shot performance highlight their potential as universal backbones for various speech processing problems, including speech emotion recognition (SER) \cite{bellver24_odyssey}.

SER is essential in applications ranging from healthcare and mental health monitoring to empathetic virtual assistants and call center analytics \cite{10152117, majidi2023utilizing}. 
However, real-world deployment of SER systems remains challenging because performance often degrades under domain mismatch, especially when training and deployment differ in corpus, speakers, channel/noise, or language \cite{cross_corpus_ser_1, ssada_cross_corpus}.

In practice, two constraints frequently arise: (i) the \emph{source-domain data} used to train specialized SER models is \emph{unavailable} at adaptation time due to privacy and ownership restrictions \cite{dataset_distillation}, and (ii) the \emph{target domain} we need to serve is \emph{unlabeled}. This combination yields the \textbf{source-free unsupervised domain adaptation (SFUDA)} setting: we must adapt to the target domain using only an already-trained source model and unlabeled target audio.

Modern deployments pose an even harder challenge. State-of-the-art LALMs, such as Gemini, are closed-source and accessible only through an API, preventing fine-tuning or inspection of their parameters. We therefore study a \textbf{harder, practical SFUDA protocol} in which the \emph{source model is a black-box LALM}. Concretely, we \emph{use the LALM as the source model for SFUDA training} and can query it, but cannot fine-tune or inspect its weights. Beyond LALMs, practitioners also often have \emph{domain-specific teacher classifiers} trained on different corpora, which they wish to evaluate on a new target domain. Our goal is not just to transfer knowledge, but to train a student that surpasses the LALM and the domain-specific teacher in the target domain. This motivates us to the central research question: \emph{given only unlabeled target-domain audio and an API-only LALM, can we adapt a student model for SER that outperforms the LALM in the target domain?}  
To answer this question, we propose \textbf{MI-Fuse}, a \textbf{denoised label fusion framework} that supplements the LALM with an additional source-domain trained SER classifier acting as an auxiliary teacher.  
Specifically, we generate multiple stochastic predictions from each teacher and compute their mean distributions. We then merge them using a weighted rule based on mutual information, where lower uncertainty across multiple generations receives a higher weight. 
To stabilize training, the classifier teacher is further updated via an exponential moving average (EMA) of the student model, ensuring that supervision evolves smoothly across iterations. 

In summary, the main contributions of this work are as follows:
\begin{itemize}[topsep=2pt, itemsep=1pt, parsep=0pt, partopsep=0pt, leftmargin=1.2em, labelsep=0.6em]
    \item We formalize a realistic and harder SFUDA scenario for speech in which the source model is a closed-source, API-only LALM, aligning with real deployment constraints.
    \item We propose a denoised label fusion approach that integrates mutual-information weighting and an EMA-updated classifier teacher to mitigate noisy pseudo-labels during adaptation.
    \item We conduct extensive experiments on multiple SER datasets, demonstrating that our approach consistently outperforms existing SFUDA methods and achieves better adaptation performance.
\end{itemize}
\vspace{-10pt}
\section{Related Work}
\vspace{-5pt}
\label{sec:related_works}
\vspace{-4pt}

To adapt a source-domain trained model to an unlabeled target domain, most SFUDA methods take inspiration from the semi-supervised learning (SSL) field and employ some common SSL techniques during adaptation to make full use of the unlabeled target data and conquer domain mismatch at the same time. These techniques could be broadly categorized into pseudo-labeling, consistency regularization, and clustering-based training. 

Common strategies for generating pseudo-labels to guide training involve class centroid-clustering \cite{liang2020we, liang2021source}, neighborhood aggregation (affinity) \cite{yang2021exploiting}, label ensembling from more than one model \cite{karim2023c, tarvainen2017mean, liu2021source}, and some even integrating complementary labels \cite{wang2024confidence} into the process. Further, consistency regularization aims to improve model robustness by enforcing consistent network predictions under either data or model variations \cite{lee2023feature}, which may also serve as another pseudo-labeling source under some settings. Besides pseudo-labeling signals (instruction), many works also integrate entropy minimization or information maximization \cite{krause2010discriminative} into the process to reduce the uncertainty of network predictions, and further promote clustering among the target features based on the clustering assumptions \cite{yang2022survey} in SSL.

Unlike prior SFUDA works that rely solely on traditional SSL or self-training techniques, our method introduces general-domain guidance by leveraging LALMs during adaptation. Specifically, we fuse pseudo-labels from a general-domain LALM with those from a source-domain teacher, enabling complementary supervision for more robust target-domain adaptation under source-free constraints.

\vspace{-8pt}
\section{Methodology}
\label{sec:method}
\vspace{-5pt}
\begin{figure}[t]
  \centering
  \includegraphics[width=0.95\columnwidth]{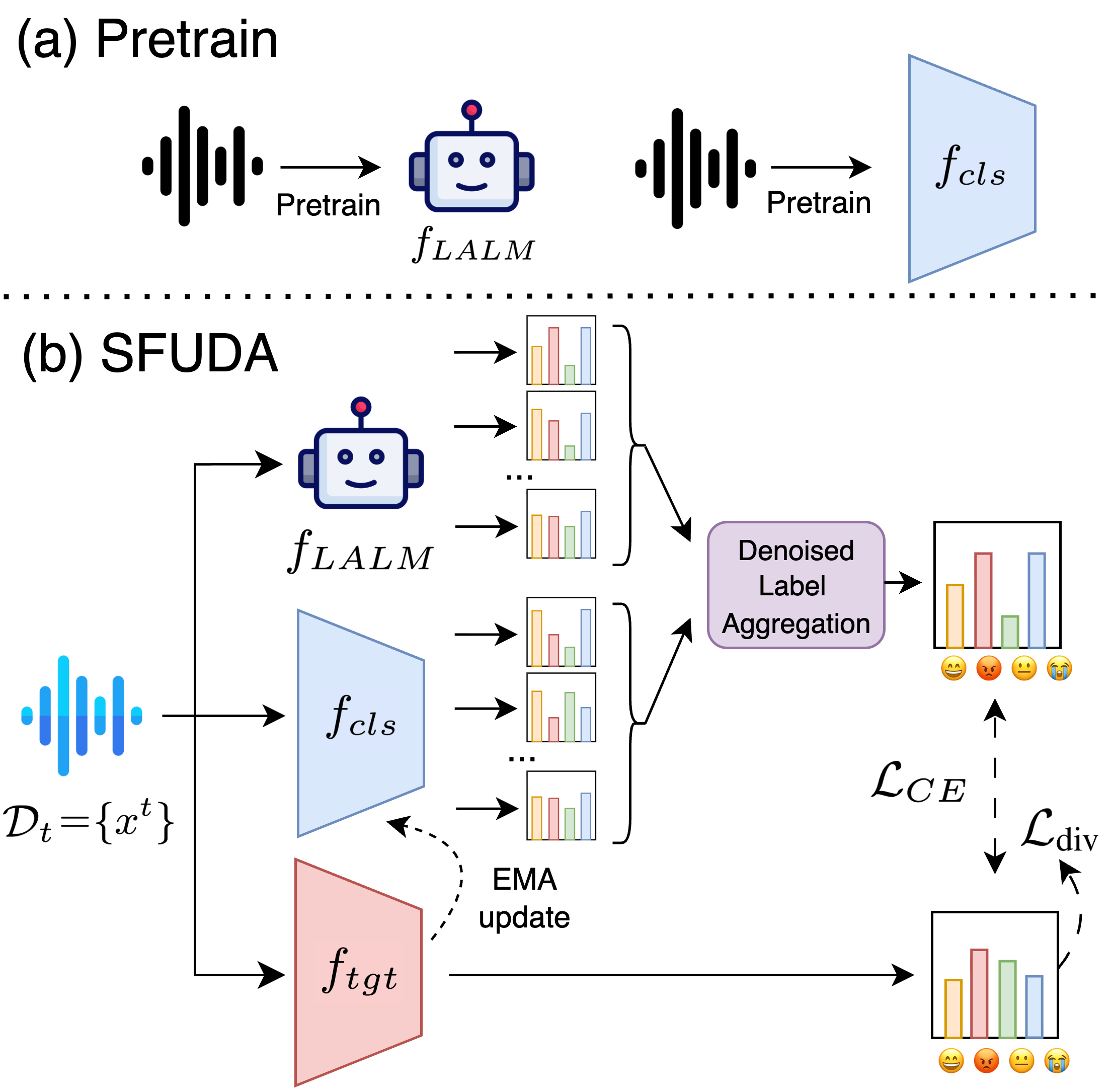}
  \vspace{-8pt}
  \caption{Overview of the SFUDA setting and our proposed method.  
  }
  \vspace{-14pt}
  \label{fig:pipeline}
\end{figure}
\subsection{Problem formulation}
\label{sec:problem_formulation}
\vspace{-3pt}
We study SFUDA for SER. The overall workflow is illustrated in Fig.~\ref{fig:pipeline}. Suppose we have a labeled source-domain dataset $\mathcal{D}_s = \{(x^s, y^s)\}$, where $x^s$ is a speech utterance, $y^s$ is its emotion label. The set of all possible labels is denoted by $\mathcal{Y} = \{1, ..., C\}$, where $C$ is the number of emotion classes. A classifier $f_{cls}$ is first trained on $\mathcal{D}_s$ and captures domain-specific SER knowledge.  

In the SFUDA setting, the original source dataset $\mathcal{D}_s$ is no longer accessible due to privacy or storage restrictions. We may still keep the trained classifier $f_{cls}$, but its performance often degrades under domain shift. In addition, we assume access to a closed-source LALM, denoted $f_{LALM}$, which is available only through an API. While $f_{LALM}$ offers strong generalization, its predictions can be noisy and cannot be fine-tuned or inspected internally.  


Finally, we are given an unlabeled target-domain dataset $\mathcal{D}_t = \{x^t\}$. The target distribution is different from the source distribution because of mismatched corpora, speakers, recording channels, or languages. The labels $y^t$ are unknown, and the goal is to adapt a student model $f_{tgt}$ that performs well on the target domain without using any labeled target data.  

A major challenge is that adaptation typically relies on pseudo-labels predicted by the source model, which can be noisy under domain shift. To overcome this, we combine the predictions of the LALM $f_{LALM}$ and the auxiliary classifier $f_{cls}$ into a denoised label distribution. This fused supervision provides a more reliable training signal for the student model $f_{tgt}$, enabling it to generalize more effectively to the target domain.
\vspace{-8pt}
\subsection{Model uncertainty estimation}
\label{sec:uncertainty}
\vspace{-4pt}
In our adaptation framework, pseudo-labels are obtained from two teachers: the $f_{LALM}$ and $f_{cls}$. Under domain shift, their predictions may be noisy, and directly trusting them risks propagating errors to the student $f_{tgt}$. We address this by explicitly estimating each teacher's uncertainty and using it in label fusion.  

We quantify uncertainty using mutual information (MI) between the predicted label $Y$ and the model parameters $\Theta$ (either $f_{LALM}$ or $f_{cls}$), conditioned on an input $x$. MI reflects how much predictions vary across stochastic perturbations such as dropout.  

For each input $x$, we perform $K$ stochastic forward passes through a teacher model ($f_{LALM}$ or $f_{cls}$). This produces a set of predictive distributions $\{p_k(y|x)\}_{k=1}^K$ (detailed in Sec.~\ref{ssec:label_fusion}). The average distribution is 
\vspace{-6pt}
\begin{equation}
\bar{p}(y|x) = \tfrac{1}{K} \sum_{k=1}^K p_k(y|x).
\vspace{-6pt}
\end{equation}
From this, the predictive entropy  
\vspace{-6pt}
\begin{equation}
H(\bar{p}) = -\sum_y \bar{p}(y|x)\,\log \bar{p}(y|x)
\vspace{-6pt}
\end{equation}
captures the total uncertainty, while the expected entropy
\vspace{-5pt}
\begin{equation}
\frac{1}{K}\sum_{k=1}^K H(p_k)
\vspace{-5pt}
\end{equation}
reflects only aleatoric uncertainty, i.e., ambiguity inherent in the input signal. Their difference, 
\vspace{-6pt}
\begin{equation}
I(Y,\Theta \mid x) = H(\bar{p}) - \tfrac{1}{K}\sum_{k=1}^K H(p_k),
\vspace{-5pt}
\end{equation}
is the mutual information (MI), which captures epistemic uncertainty as the model’s disagreement across stochastic predictions.

This measure is directly applicable to SFUDA. A high MI value means the teacher’s predictions vary significantly across samples, signaling that its output is unstable under domain shift and should be down-weighted. A low MI value indicates stable and consistent predictions that are more trustworthy. We weight each teacher’s contribution by $e^{-MI}$ to suppress noise and amplify consistency.  

Integrating MI into our label fusion mechanism provides the student model $f_{tgt}$ with less noisy and more robust pseudo-labels. This allows $f_{tgt}$ to benefit from both the generalization ability of the LALM and the domain-specific knowledge of the classifier, while mitigating the risk of propagating errors from either teacher.  

\vspace{-6pt}
\subsection{Label fusion}
\vspace{-2pt}
\label{ssec:label_fusion}
We propose a novel label fusion framework for SFUDA in SER. Our approach integrates predictions from a source-pretrained SER model and an LALM to mitigate the noisy label problem. 
We begin by initializing the student model $f_{tgt}$ with the parameters of the classifier teacher $f_{cls}$. The student is then adapted using only unlabeled target domain data. 

For each unlabeled target sample $x_t$, we apply \textit{Monte Carlo (MC) dropout} \cite{mc_dropout} to the classifier teacher and obtain $N_{cls}$ stochastic forward passes. This produces a set of predicted probability distributions $$
\{ p^{(1)}_{\text{cls}}(y|x_t), p^{(2)}_{\text{cls}}(y|x_t), \dots, p^{(N)}_{\text{cls}}(y|x_t) \}.
$$ The mean distribution is used as the aggregated classifier teacher prediction:
\vspace{-4pt}
\begin{equation}
\bar{p}_{\text{cls}}(y|x_t) = \frac{1}{N_{cls}} \sum_{i=1}^{N_{cls}} p^{(i)}_{\text{cls}}(y|x_t).
\vspace{-4pt}
\end{equation}

In parallel, we query the LALM $N_{LM}$ times with carefully designed prompts that request a probability distribution over emotion classes in natural language. This yields 
\vspace{-4pt}
$$
\{ p^{(1)}_{\text{LM}}(y|x_t), p^{(2)}_{\text{LM}}(y|x_t), \dots, p^{(N)}_{\text{LM}}(y|x_t) \}.\vspace{-4pt}$$ We then compute the mean LALM probability distribution:
\vspace{-6pt}
\begin{equation}
\bar{p}_{\text{LM}}(y|x_t) = \frac{1}{{N_{LM}}} \sum_{i=1}^{N_{LM}} p^{(i)}_{\text{LM}}(y|x_t).
\vspace{-6pt}
\end{equation}

We fuse the outputs from two models by weighted averaging. The weights are determined by the exponential of the negative mutual information (MI) associated with each model’s distribution,
\vspace{-4pt}
\begin{equation}
p_{\text{fused}}(y|x_t) = \frac{e^{-\text{MI}_1} \cdot \bar{p}_{\text{cls}}(y|x_t) + e^{-\text{MI}_2} \cdot \bar{p}_{\text{LM}}(y|x_t)}{e^{-\text{MI}_1} + e^{-\text{MI}_2}},
\vspace{-4pt}
\end{equation}
thereby assigning greater importance to the model with lower uncertainty. We set ${N_{LM}}=5$ and ${N_{cls}}=8$ passes in our experiments.
\vspace{-8pt}
\subsection{Diversity Loss and EMA Teacher Update}
\label{sec:losses}

The student model $f_{tgt}$ is trained on the fused pseudo-labels by minimizing a cross-entropy loss with soft targets. However, relying solely on pseudo-label supervision risks two common issues in SFUDA: (1) \textit{class collapse}, where the model overfits to a subset of emotions while ignoring others, and (2) \textit{unstable supervision}, where the teacher’s predictions degrade over time due to noisy labels and representation drift. To address these issues, we introduce two complementary mechanisms: a diversity loss and an exponential moving average (EMA) teacher update.  


\textbf{Diversity loss.}  
To prevent class collapse, we encourage the model to maintain high entropy across predictions at the batch level, following \cite{pmlr-v119-liang20a}. Specifically, given a batch of student predicted probability distributions $\{p_{\text{tgt}}(y|x_t^i)\}_{i=1}^B$, we compute the average prediction:  
\vspace{-5pt}
\begin{equation}
\bar{p}_{\text{batch}}(y) = \frac{1}{B}\sum_{i=1}^B p_{\text{tgt}}(y|x_t^i).
\vspace{-5pt}
\end{equation}
The diversity loss is then defined as $\mathcal{L}_{\text{div}} = -H(\bar{p}_{\text{batch}})$, which pushes the model to spread its predictions more evenly across emotion classes. In practice, this promotes balanced learning and reduces the effect of noisy pseudo-labels.  

\textbf{EMA teacher update.}  
To mitigate unstable supervision, we update the classifier teacher $f_{cls}$ as an exponential moving average of the student $f_{tgt}$. At each step, the teacher parameters $\theta_{cls}$ are updated by  
\vspace{-1pt}
\begin{equation}
\theta_{cls} \leftarrow \alpha \,\theta_{cls} + (1-\alpha)\,\theta_{tgt},
\end{equation}
where $\theta_{tgt}$ are the student parameters and $\alpha \in [0,1)$ is a momentum factor. This ensures that the teacher evolves smoothly with the student, filtering out short-term noise and providing more stable guidance during training. We use $\alpha=0.999$.


\textbf{Overall objective.}  
The final training objective of the student model combines supervised alignment with regularization:  
\vspace{-4pt}
\begin{equation}
\mathcal{L} = \mathcal{L}_{\text{CE}} + \lambda_{\text{div}} \mathcal{L}_{\text{div}},
\vspace{-4pt}
\end{equation}
where $\mathcal{L}_{\text{CE}}$ is the cross-entropy between the student predictions and the fused pseudo-labels, and $\lambda_{\text{div}}$ are hyperparameters balancing the contributions of the two regularizers. We set $\lambda_{\text{div}}=1$. 

\begin{table*}[tbp]
\centering
\caption{Performance across dataset transfer tasks in Accuracy (\%). Best results are \textbf{bolded} and second-best are \underline{underlined}.}
\vspace{1pt}
\begin{adjustbox}{max width=\textwidth}
\begin{tabular}{lccccccc}  
\toprule
\textbf{Method}
& \textbf{IMP $\rightarrow$ POD}
& \textbf{POD $\rightarrow$ IMP}
& \textbf{IEM $\rightarrow$ IMP}
& \textbf{IMP $\rightarrow$ IEM}
& \textbf{POD $\rightarrow$ IEM}
& \textbf{IEM $\rightarrow$ POD}
& \textbf{Avg.} \\   
\midrule
LALM SFUDA              & 60.59 & \second{56.74} & 51.75 & 48.40 & 51.27 & 58.12 & \second{54.48} \\
LALM zero-shot          & \second{61.44} & 53.66 & \second{53.66} & 45.96 & 45.96 & \best{61.44} & 53.69 \\
Source model SFUDA     & 41.34 & \second{56.74} & 51.48 & \second{53.75} & 53.85 & 48.90 & 51.01 \\
Source model zero-shot & 41.37 & \second{56.74} & 50.50 & 49.08 & 41.27 & 48.90 & 49.64 \\
\midrule
SHOT                   & 41.58 & 56.51 & 50.64 & 50.13 & 55.94 & 48.90 & 50.62 \\
NRC                    & 41.37 & \second{56.74} & 50.48 & 52.09 & \best{59.61} & 48.90 & 51.53 \\
\midrule
\textbf{MI-Fuse}          & \best{61.92} & \best{57.48} & \best{54.87} & \best{59.09} & \second{57.07} & \second{59.85} & \best{58.38} \\
\bottomrule
\end{tabular}
\end{adjustbox}
\label{tab:overall_6_tasks}
\vspace{-14pt}
\end{table*}
\vspace{-7pt}
\section{Experiments}
\vspace{-4pt}
\subsection{Setup}
\vspace{-4pt}
\noindent\textbf{Dataset.} This study uses three publicly available emotion databases, MSP-Podcast\cite{msp_podcast}, IMPROV\cite{msp_improv}, and IEMOCAP\cite{iemocap}, which are denoted as POD, IMP, and IEM, respectively. They cover real-world and acted emotions across different ethnic groups, ensuring diversity. Both IMPROV and IEMOCAP undergo cross-validation settings, including six and five folds for thorough evaluation, while there's only one fold in Podcast. To perform cross-dataset learning, we further filter the datasets to 4 emotion categories (happy, sad, angry, and neutral). We use unweighted accuracy as the evaluation metric.

\noindent\textbf{Model.}
We use Gemini 2.5 flash as the LALM studied, with a temperature of 0.6 for text generation. For the classifier teacher and student, we use the WavLM base+ model \cite{wavlm} with a weighted sum across layers for feature extraction. The extracted representations are then passed through two linear layers to predict the final emotion category, following \cite{emobias, emodebias, lin25c_interspeech}. The classifier teacher is trained on the source domain using a cross-entropy loss $\mathcal{L}_{CE}$. 

\noindent\textbf{Optimization.} We optimize the networks using the AdamW \cite{adamw} optimizer, with a batch size of 32. For regularization, we apply dropout with a rate of 0.4 in the linear layers, and an $L_2$ regularization with weight 0.1. The teacher classifier $f_{cls}$ is trained with a learning rate of 5e-4, and the student model $f_{tgt}$ is scanned through learning rates of \{7.5e-4, 5e-4, 1e-4, 5e-5, 1e-6\}. The models are trained until the loss stops decreasing for 1000 steps. 
\vspace{-7pt}
\subsection{Compare with other methods}
\vspace{-3pt}
We compare six baselines spanning zero-shot inference, single-teacher target adaptation, and standard SFUDA. No baseline uses source-domain audio.

\begin{itemize} [topsep=2pt, itemsep=1pt, parsep=0pt, partopsep=0pt, leftmargin=1.2em, labelsep=0.6em]
  \item \textbf{LLM zero-shot}: Query the LALM $f_{\text{LALM}}$ for class probabilities and take argmax. No target training.
  \item \textbf{Source model zero-shot}: Evaluate the domain-specific classifier $f_{\text{cls}}$ directly on target utterances. No target training.
  \item \textbf{LALM / Source model SFUDA}: Adapt a student $f_{tgt}$ on unlabeled target data using pseudo-labels from $f_{\text{LALM}}$/$f_{\text{cls}}$ only; the other teacher is unused.
  \item \textbf{SHOT \cite{liang2020we}/ NRC \cite{yang2021exploiting}}: State-of-the-art SFUDA methods implemented with our backbone using $f_{\text{cls}}$ teacher.
\end{itemize}

\noindent\textbf{Overall performance.}  
Our denoised label fusion framework consistently outperforms all baselines. Averaged across six transfer settings, MI-Fuse achieves 58.38\% unweighted accuracy, which is 3.9\% higher than the best-performing baseline (LALM SFUDA). This demonstrates that combining the API-based LALM with a source-trained classifier through denoised label fusion provides more reliable supervision than relying on either teacher alone.  

\noindent\textbf{Performance on individual transfers.}  
MI-Fuse achieves the highest accuracy in four of the six transfer directions. For example, in the IMP $\rightarrow$ POD setting, the LALM zero-shot baseline already performs strongly (61.44\%), but MI-Fuse further improves to 61.92\%. In the IMP $\rightarrow$ IEM setting, where the source model performs better than the LALM (53.75\% vs. 45.96\%), our framework effectively integrates both sources of information and raises performance to 59.09\%, far exceeding the strongest SFUDA baselines SHOT (50.13\%) and NRC (52.09\%). These results highlight the advantage of uncertainty-aware label fusion when teacher reliability varies across tasks.  

\noindent\textbf{Competitive performance in challenging cases.}  
Even in the two settings where MI-Fuse does not achieve the top score, it still ranks second. For instance, in IEM $\rightarrow$ POD, our accuracy of 59.85\% is close to the LALM zero-shot performance of 61.44\%. Importantly, while conventional SFUDA methods such as SHOT and NRC are sometimes competitive when adapting to the IEMOCAP corpus, they lag behind in other domains. Our approach maintains consistently high performance across all transfer directions, demonstrating stronger generalization.  
\vspace{-13pt}
\subsection{Ablation study}
\vspace{-3pt}
To validate the effectiveness of our label fusion framework, we implement plausible alternatives for each component. \textbf{(i) Generation strategy:} 
\emph{Multi} uses $N_{LM}=5, N_{cls}=8$ stochastic passes for teachers (temperature sampling for $f_{\text{LALM}}$ and Monte Carlo dropout for $f_{\text{cls}}$) and averages the resulting distributions; \emph{Single} disables stochasticity ($N_{LM}{=}1$) for $f_{\text{LALM}}$ using zero temperature. 
\textbf{(ii) \textbf{Similarity gate}:} 
\emph{Direct} fusion always fuses the two mean distributions; \emph{KL} fuses only when $D_{\mathrm{KL}}(\bar p_{\text{cls}}\Vert \bar p_{\text{LM}})\le\tau$ (with $\tau$ grid-searched from $\{0.4, 0.6, 0.8\}$), otherwise it selects the lower-entropy teacher; \emph{No Fusion} always uses the labels from the teacher with lowest entropy.
\textbf{(iii) Weighting:} when fusing, we combine soft labels from teachers using one of: \emph{MI} (weights $\propto e^{-\mathrm{MI}}$ per teacher), \emph{Entropy} (weights $\propto e^{-H}$), or \emph{Equal} (uniform average).


Table~\ref{tab:ablation} shows that our full method MI-Fuse (\emph{Multi + Direct Fusion + MI}) achieves the best accuracy on both transfer directions.
KL gating consistently underperforms direct fusion across weightings, indicating that hard disagreement gating discards useful complementary cues. Single-teacher training is clearly weaker, and entropy-based weighting lags behind MI, confirming the benefit of epistemic-uncertainty aware fusion.

\begin{table}[t]   
  \centering
  \caption{Ablation study on fusion strategies on IEMOCAP (Accuracy\,\%). Best performances are in \textbf{bold}.}
  \vspace{2pt}
  \renewcommand{\arraystretch}{1.15}

  \begin{adjustbox}{max width=\columnwidth}
    \begin{tabular}{lllcc}
      \toprule
      \cmidrule(lr){4-5}
      \textbf{Generation} & \textbf{Similarity} & \textbf{Weighting} &
      \textbf{IMP $\rightarrow$ IEM} & \textbf{POD $\rightarrow$ IEM}\\
      \midrule
      \multirow{7}{*}{Multi}
        & \multirow{3}{*}{Direct}
          & MI (Ours) & \best{59.09} & \best{57.07} \\
        &          & Entropy& 57.34 & 55.53 \\
        &          & Equal  & 57.98 & \second{56.64} \\ \cmidrule{2-5}
        & \multirow{3}{*}{KL}
          & MI      & 55.23 & 55.86 \\
        &          & Entropy& 55.13 & 55.09 \\
        &          & Equal  & 55.17 & 55.12 \\ \cmidrule{2-5}
        & No Fusion & -     & 56.08 & 55.43 \\
      \midrule
      \multirow{5}{*}{Single}
        & \multirow{2}{*}{Direct}
          & Entropy & 56.82 & 55.85 \\
        &          & Equal  & \second{58.26} & 55.23 \\ \cmidrule{2-5}
        & \multirow{2}{*}{KL}
          & Entropy & 54.05 & 51.71 \\
        &          & Equal  & 54.16 & 51.79 \\ \cmidrule{2-5}
        & No Fusion & -     & 54.05 & 51.69 \\
      \bottomrule
    \end{tabular}
  \end{adjustbox}
  \vspace{-15pt}
  \label{tab:ablation}
\end{table}
\vspace{-7pt}
\subsection{Training stability analysis}
\begin{figure}[tbp]
  \centering
  \includegraphics[width=0.40\textwidth]{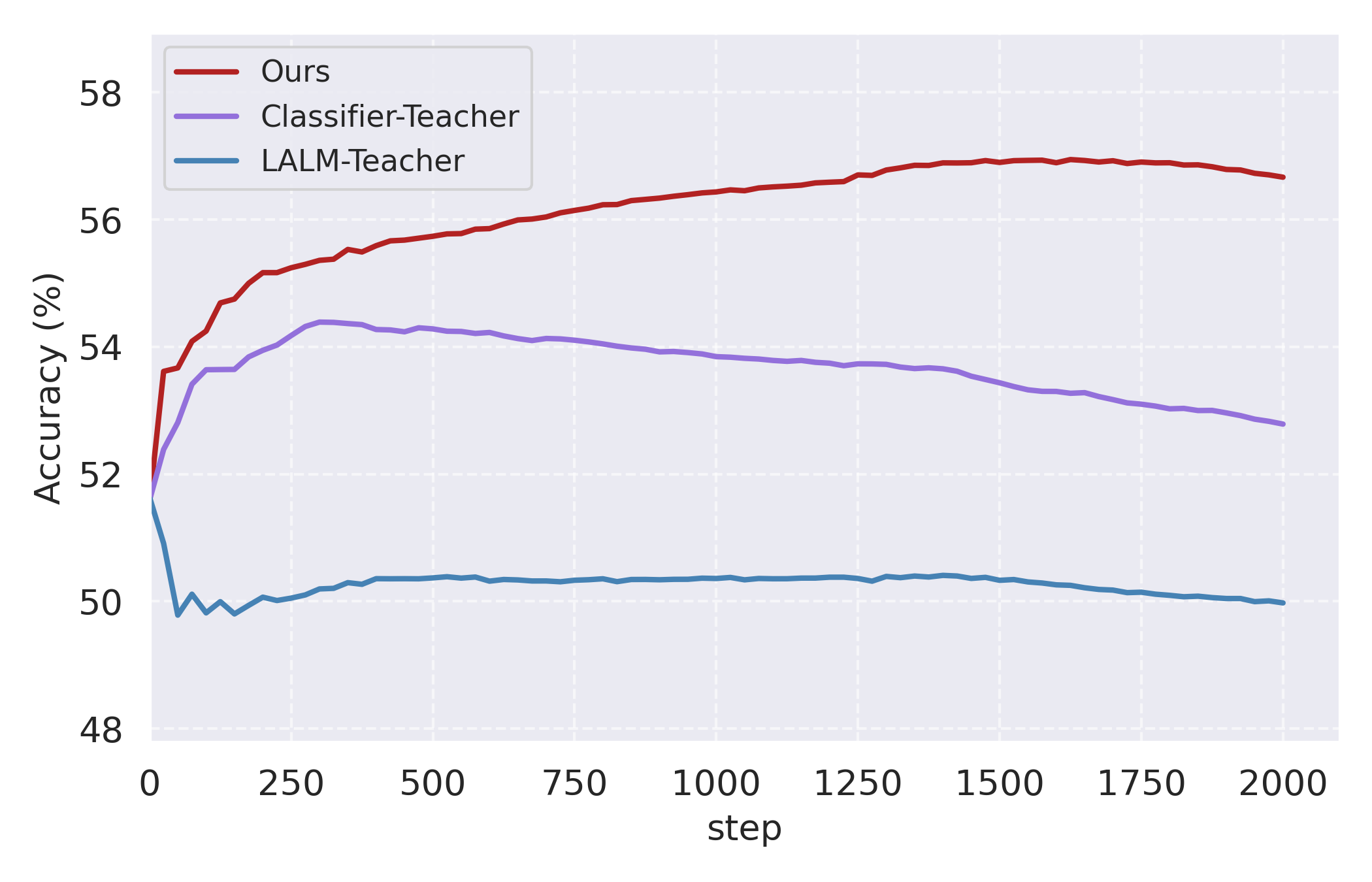}
  \vspace{-10pt}
  \caption{Development set accuracy over training steps on IMP $\rightarrow$ IEM fold 1. Our method (red) consistently outperforms both baselines: the classifier teacher (purple) and the LALM teacher (blue).}
  \label{fig:acc-propagation}
  \vspace{-10pt}
\end{figure}

\vspace{-3pt}
Fig.~\ref{fig:acc-propagation} shows how the development set accuracy evolves during the adaptation, revealing that our proposed approach achieves not only higher final accuracy but also more stable training dynamics compared to the classifier teacher and LALM teacher baselines. The classifier teacher declines after $\sim$400 steps, due to overfitting on early pseudo-labels from the EMA teacher. The LALM teacher performs the worst, dropping sharply at the start and then stagnating, which reflects unreliable predictions under domain shift. In contrast, our method steadily improves throughout training, effectively balancing information from both teachers while suppressing noise.  

\vspace{-8pt}
\section{Limitation}
\label{sec:limitation}
\vspace{-3pt}
First, MI-Fuse depends on LALMs that can produce meaningful probabilistic predictions over emotion categories. Although models such as Gemini 2.5 are becoming increasingly accessible, their inference cost, latency, and reliance on proprietary APIs may hinder practical deployment in resource-constrained or privacy-sensitive settings. Second, the label fusion scheme assumes a fixed set of discrete emotion categories across datasets. In real-world applications, however, emotion taxonomies may vary. This mismatch can hinder the direct applicability of MI-Fuse when adapting to target domains with different label spaces. 

\vspace{-3pt}
\section{Conclusion}
\label{sec:conclusion}
\vspace{-2pt}
We present MI-Fuse, a denoised label fusion framework for SFUDA in SER under realistic closed-source LALM constraints. By integrating mutual-information-aware fusion, a diversity loss, and an EMA-updated teacher, our approach produces stable pseudo-labels. It effectively balances the strengths of both general-purpose LALMs and domain-trained classifiers. Extensive experiments across multiple datasets demonstrate that MI-Fuse consistently improves cross-domain performance, surpasses strong baselines, and enables student models to outperform closed-source LALMs on target domains. These results establish MI-Fuse as a practical recipe for deploying emotion-aware speech systems under realistic constraints.

\bibliographystyle{IEEEbib}
\bibliography{strings,refs}

\begin{thebibliography}{10}

\bibitem{desta2.5_audio}
Ke-Han Lu et~al.,
\newblock ``Desta2. 5-audio: Toward general-purpose large audio language model with self-generated cross-modal alignment,''
\newblock {\em arXiv preprint arXiv:2507.02768}, 2025.

\bibitem{qwen2.5_omni}
Jin Xu et~al.,
\newblock ``Qwen2. 5-omni technical report,''
\newblock {\em arXiv preprint arXiv:2503.20215}, 2025.

\bibitem{gemini2.5}
Gheorghe Comanici et~al.,
\newblock ``Gemini 2.5: Pushing the frontier with advanced reasoning, multimodality, long context, and next generation agentic capabilities,''
\newblock {\em arXiv preprint arXiv:2507.06261}, 2025.

\bibitem{dynamic_superb_phase_2}
Chien yu~Huang et~al.,
\newblock ``Dynamic-{SUPERB} phase-2: A collaboratively expanding benchmark for measuring the capabilities of spoken language models with 180 tasks,''
\newblock in {\em The Thirteenth International Conference on Learning Representations}, 2025.

\bibitem{wu2024towards}
Haibin Wu et~al.,
\newblock ``Towards audio language modeling--an overview,''
\newblock {\em arXiv preprint arXiv:2402.13236}, 2024.

\bibitem{bellver24_odyssey}
Jaime Bellver et~al.,
\newblock ``Multimodal audio-language model for speech emotion recognition,''
\newblock in {\em The Speaker and Language Recognition Workshop (Odyssey 2024)}, 2024, pp. 288--295.

\bibitem{10152117}
Nelly Elsayed et~al.,
\newblock ``Speech emotion recognition using supervised deep recurrent system for mental health monitoring,''
\newblock in {\em 2022 IEEE 8th World Forum on Internet of Things (WF-IoT)}, 2022.

\bibitem{majidi2023utilizing}
Farideh Majidi and Marzieh Bahrami,
\newblock ``Utilizing speech emotion recognition and recommender systems for negative emotion handling in therapy chatbots,''
\newblock {\em arXiv preprint arXiv:2311.11116}, 2023.

\bibitem{cross_corpus_ser_1}
Cheng Lu et~al.,
\newblock ``Progressively discriminative transfer network for cross-corpus speech emotion recognition,''
\newblock {\em Entropy}, vol. 24, 2022.

\bibitem{ssada_cross_corpus}
Siddique Latif et~al.,
\newblock ``Self supervised adversarial domain adaptation for cross-corpus and cross-language speech emotion recognition,''
\newblock {\em IEEE Transactions on Affective Computing}, vol. 14, no. 3, pp. 1912--1926, 2023.

\bibitem{dataset_distillation}
Fabian Ritter-Gutierrez et~al.,
\newblock ``{Dataset-Distillation Generative Model for Speech Emotion Recognition},''
\newblock in {\em {Interspeech 2024}}, 2024, pp. 2640--2644.

\bibitem{liang2020we}
Jian Liang, Dapeng Hu, and Jiashi Feng,
\newblock ``Do we really need to access the source data? source hypothesis transfer for unsupervised domain adaptation,''
\newblock in {\em International conference on machine learning}. PMLR, 2020, pp. 6028--6039.

\bibitem{liang2021source}
Jian Liang et~al.,
\newblock ``Source data-absent unsupervised domain adaptation through hypothesis transfer and labeling transfer,''
\newblock {\em IEEE Transactions on Pattern Analysis and Machine Intelligence}, vol. 44, no. 11, pp. 8602--8617, 2021.

\bibitem{yang2021exploiting}
Shiqi Yang et~al.,
\newblock ``Exploiting the intrinsic neighborhood structure for source-free domain adaptation,''
\newblock {\em Advances in neural information processing systems}, vol. 34, pp. 29393--29405, 2021.

\bibitem{karim2023c}
Nazmul Karim et~al.,
\newblock ``C-sfda: A curriculum learning aided self-training framework for efficient source free domain adaptation,''
\newblock in {\em Proceedings of the IEEE/CVF conference on computer vision and pattern recognition}, 2023, pp. 24120--24131.

\bibitem{tarvainen2017mean}
Antti Tarvainen and Harri Valpola,
\newblock ``Mean teachers are better role models: Weight-averaged consistency targets improve semi-supervised deep learning results,''
\newblock {\em Advances in neural information processing systems}, vol. 30, 2017.

\bibitem{liu2021source}
Yuang Liu et~al.,
\newblock ``Source-free domain adaptation for semantic segmentation,''
\newblock in {\em Proceedings of the IEEE/CVF conference on computer vision and pattern recognition}, 2021, pp. 1215--1224.

\bibitem{wang2024confidence}
Jincen Wang et~al.,
\newblock ``Confidence-aware hypothesis transfer networks for source-free cross-corpus speech emotion recognition,''
\newblock in {\em Proc. Interspeech 2024}, 2024, pp. 1050--1054.

\bibitem{lee2023feature}
JoonHo Lee and Gyemin Lee,
\newblock ``Feature alignment by uncertainty and self-training for source-free unsupervised domain adaptation,''
\newblock {\em Neural Networks}, vol. 161, pp. 682--692, 2023.

\bibitem{krause2010discriminative}
Andreas Krause et~al.,
\newblock ``Discriminative clustering by regularized information maximization,''
\newblock {\em Advances in neural information processing systems}, vol. 23, 2010.

\bibitem{yang2022survey}
Xiangli Yang et~al.,
\newblock ``A survey on deep semi-supervised learning,''
\newblock {\em IEEE transactions on knowledge and data engineering}, vol. 35, no. 9, pp. 8934--8954, 2022.

\bibitem{mc_dropout}
Yarin Gal and Zoubin Ghahramani,
\newblock ``Dropout as a bayesian approximation: Representing model uncertainty in deep learning,''
\newblock in {\em Proceedings of The 33rd International Conference on Machine Learning}. 2016, vol.~48 of {\em Proceedings of Machine Learning Research}, pp. 1050--1059, PMLR.

\bibitem{pmlr-v119-liang20a}
Jian Liang et~al.,
\newblock ``Do we really need to access the source data? {S}ource hypothesis transfer for unsupervised domain adaptation,''
\newblock in {\em Proceedings of the 37th International Conference on Machine Learning}, 2020.

\bibitem{msp_podcast}
Reza Lotfian and Carlos Busso,
\newblock ``Building naturalistic emotionally balanced speech corpus by retrieving emotional speech from existing podcast recordings,''
\newblock {\em IEEE Transactions on Affective Computing}, vol. 10, no. 4, pp. 471--483, 2019.

\bibitem{msp_improv}
Carlos Busso, Srinivas Parthasarathy, Alec Burmania, Mohammed AbdelWahab, Najmeh Sadoughi, and Emily~Mower Provost,
\newblock ``Msp-improv: An acted corpus of dyadic interactions to study emotion perception,''
\newblock {\em IEEE Transactions on Affective Computing}, vol. 8, no. 1, pp. 67--80, 2017.

\bibitem{iemocap}
Carlos Busso et~al.,
\newblock ``Iemocap: Interactive emotional dyadic motion capture database,''
\newblock {\em Language resources and evaluation}, 2008.

\bibitem{wavlm}
Sanyuan Chen et~al.,
\newblock ``Wavlm: Large-scale self-supervised pre-training for full stack speech processing,''
\newblock {\em IEEE Journal of Selected Topics in Signal Processing}, 2022.

\bibitem{emobias}
Yi-Cheng Lin et~al.,
\newblock ``{Emo-bias: A Large Scale Evaluation of Social Bias on Speech Emotion Recognition},''
\newblock in {\em {Interspeech 2024}}, 2024, pp. 4633--4637.

\bibitem{emodebias}
Yi-Cheng Lin et~al.,
\newblock ``Emo-debias: Benchmarking gender debiasing techniques in multi-label speech emotion recognition,''
\newblock {\em arXiv preprint arXiv:2506.04652}, 2025.

\bibitem{lin25c_interspeech}
Yi-Cheng Lin et~al.,
\newblock ``{Mitigating Subgroup Disparities in Multi-Label Speech Emotion Recognition: A Pseudo-Labeling and Unsupervised Learning Approach},''
\newblock in {\em {Interspeech 2025}}, 2025.

\bibitem{adamw}
Ilya Loshchilov and Frank Hutter,
\newblock ``Decoupled weight decay regularization,''
\newblock in {\em International Conference on Learning Representations}, 2019.

\end{thebibliography}

\end{document}